\documentclass[10pt,twocolumn,letterpaper]{article}

\usepackage{cvpr}
\usepackage{times}
\usepackage{epsfig}
\usepackage{graphicx}
\usepackage{tabularx}
\usepackage{amsmath}
\usepackage{amssymb}
\usepackage[inline]{enumitem}
\usepackage{multirow}
\usepackage{subcaption}
\usepackage[dvipsnames]{xcolor}
\usepackage{booktabs}
\usepackage{algpseudocode,algorithm,algorithmicx}
\usepackage{xspace}

\graphicspath{{figures/}}

\newcommand{\system}{BlockDrop\xspace}
\newcommand{\ra}[1]{\renewcommand{\arraystretch}{#1}}
	
\newcommand*\Let[2]{\State #1 $\gets$ #2}
\algrenewcommand\algorithmicrequire{\textbf{Input:}}
\algrenewcommand\algorithmicensure{\textbf{Result:}}
\definecolor{citecolor}{RGB}{34, 139, 34}

\newcommand{\ZW}[1]{{\color{black}{#1}}}

\newcommand{\commentout}[1]{}

\usepackage[pagebackref=true,breaklinks=true,letterpaper=true,colorlinks,citecolor=citecolor, bookmarks=false]{hyperref}

\cvprfinalcopy %

\def\cvprPaperID{***} %

\ifcvprfinal\fi
\begin{document}

\title{BlockDrop: Dynamic Inference Paths in Residual Networks}
\newcommand\blfootnote[1]{%
  \begingroup
  \renewcommand\thefootnote{}\footnote{#1}%
  \addtocounter{footnote}{-1}%
  \endgroup
}

\author{Zuxuan Wu$^{1*}$, Tushar Nagarajan$^{2*}$, Abhishek Kumar$^{3}$, Steven Rennie$^{4}$ \\ Larry S. Davis$^{1}$, Kristen Grauman$^{2}$, Rogerio Feris$^{3}$ \\
$^{1}$ UMD, $^{2}$ UT Austin, $^{3}$ IBM Research, $^{4}$ Fusemachines Inc.}

\maketitle

\begin{abstract}
\blfootnote{$^{*}$ Authors contributed equally}
Very deep convolutional neural networks offer excellent recognition results, yet their computational expense limits their impact for many real-world applications.  We introduce BlockDrop, an approach that learns to dynamically choose which layers of a deep network to execute during inference so as to best reduce total computation without degrading prediction accuracy.   Exploiting the robustness of Residual Networks (ResNets) to layer dropping, our framework selects on-the-fly which residual blocks to evaluate for a given novel image.
In particular, given a pretrained ResNet, we train a policy network in an associative reinforcement learning setting for the dual reward of utilizing a minimal number of blocks while preserving recognition accuracy. We conduct extensive experiments on CIFAR and ImageNet.  The results provide strong quantitative and qualitative evidence that these learned policies not only accelerate inference but also encode meaningful visual information. Built upon a ResNet-101 model, our method achieves a speedup of 20\% on average, going as high as 36\% for some images, while maintaining the same 76.4\% top-1 accuracy on ImageNet.
\end{abstract}
\thispagestyle{empty}

\section{Introduction}
\label{sec:intro}

Deep neural networks are now ubiquitous in computer vision owing to their recent successes in several important tasks.   However, great strides in accuracy have been accompanied by increasingly complex and deep network architectures.  This presents a problem for domains where fast inference is essential, particularly in delay-sensitive and real-time scenarios such as autonomous driving, robotic navigation, or user-interactive applications on mobile devices.

\begin{figure}[t!]
\centering
\includegraphics[width=0.85\columnwidth]{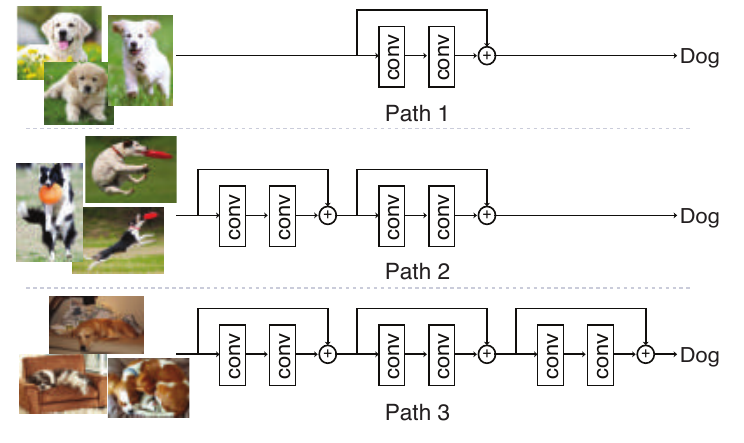}
\vspace{-0.1in}
\caption{ \textbf{A conceptual overview of our approach}. Rather than execute all blocks of a ResNet, our approach learns a policy to select the minimal configuration of blocks that is needed to correctly classify a given input image.  The resulting instance-specific paths in the network not only reflect the image's difficulty (easier samples use fewer blocks) but also encode meaningful visual information (patterns of blocks correspond to clusters of visual features).} 
\label{fig:teaser}
\vspace{-0.2in}
\end{figure}

Most existing work pursues model compression techniques to speed up a deep network~\cite{Hinton15,chenNIPS2017,ioannou2015training,sainath2013low,polyak2015channel,li2016pruning,han2015deep,wu2016quantized,li2017training}.  While significant speed-ups are possible, the approach yields a one-size-fits-all network that requires the same fixed set of features to be extracted for all novel images, no matter their  complexity.  In contrast, an important feature of the human perception system is its ability to adaptively allocate time and scrutiny for visual recognition~\cite{ude2012integrating}. For example, a single glimpse is sufficient to recognize some objects and scenes, whereas more time and attention is required to clearly understand occluded or complicated ones~\cite{walther2011simple}.

In this spirit, we explore the problem of dynamically allocating computation across a deep network.  %
In particular, we consider Residual Networks (ResNet)~\cite{he2015deep} both due to their strong track record for recognition tasks~\cite{he2015deep,dai2016r,he2017mask} as well as their tolerance to removal of  layers~\cite{veit2016residual}.  
ResNets are composed of \emph{residual blocks}, consisting of two or more convolutional layers and skip-connections, which enable direct paths between any two residual blocks. 
These skip-connections %
make ResNets behave like ensembles of relatively shallow networks, and hence the removal of a certain residual block generally has only a modest impact on performance~\cite{veit2016residual}. 
However, the preliminary study of block dropping in ResNets~\cite{veit2016residual} applies a global, manually defined dropping scheme (the same blocks for all images), which leads to increased errors when more blocks are dropped. 

We propose to learn optimal block dropping strategies that simultaneously preserve both prediction accuracy and minimal block usage based on image-specific decisions. 
When a novel input is presented to the network trained for recognition, a \emph{dynamic inference path} is followed, selectively choosing which blocks to compute for that instance.  See Figure~\ref{fig:teaser}.  The approach not only improves computational efficiency during inference (\ie, for a similar prediction accuracy, being able to drop more residual blocks than a static global scheme), but also facilitates further insights into ResNets, \eg, whether different blocks encode information about objects, whether the computation needed to classify depends on the difficulty level of the example.  

To this end, we introduce \emph{\system}, a reinforcement learning approach to derive instance-specific inference paths in ResNets.  The main idea is to learn a model (referred to as the \emph{policy network}) that, given a novel input image, outputs the posterior probabilities of all the binary decisions for dropping or keeping each block in a pretrained ResNet. The policy network is trained using curriculum learning to maximize a reward that incentivizes the use of as few blocks as possible while preserving the prediction accuracy. In addition, the pretrained ResNet is further jointly finetuned with the policy network to produce feature transformations tailored for block dropping behavior.  Our approach can be seen as an instantiation of associative reinforcement learning~\cite{sutton1998reinforcement} where all the decisions are taken in a single step given the context (\ie, the input instance)\footnote{It can also be seen as contextual bandits \cite{langford2008epoch} although we do not operate in an online setting which has an objective of minimizing the \emph{regret}.}; this makes policy execution lightweight and scalable to very deep networks.  

We conduct extensive experiments on CIFAR~\cite{krizhevsky2009learning} and ImageNet~\cite{deng2009imagenet}.
\system achieves 93.6\% and 73.7\% accuracy using just 33\% and 55\% of blocks in a pretrained ResNet-110 on CIFAR-10 and CIFAR-100, respectively, outperforming state-of-the-art methods~\cite{figurnov2017spatially,graves2016adaptive,dong2017more,li2016pruning} by clear margins. Furthermore, \system speeds up a ResNet-101 model on ImageNet by 20\% while maintaining the same 76.4\% top-1 accuracy~\footnote{\url{https://goo.gl/EwHQcq}}. 
Qualitatively, we observe that the dropping policies learned with \system are correlated with the visual patterns in the images, \eg, within the ``orange'' class, images containing a pile of oranges take an inference path that is different from that taken by the close-up images of oranges. Furthermore, \system policies for \emph{easy} images with clearly visible objects utilize fewer residual blocks compared to the \emph{difficult} images that contain other occluding or background objects.  
Note that although our analysis in this paper is focused on vanilla ResNets, our approach could also be applied to other recently proposed ResNet variants such as ResNeXt \cite{xie2017aggregated} or Multi-Residual Networks \cite{abdi2016multi}, as well as other tasks beyond image classification.

\section{Related Work}
\noindent \textbf{Layer Dropping in Residual Networks}. 
Dropping layers in residual networks has been used as a regularization mechanism, similar to Dropout~\cite{srivastava2014dropout} or DropConnect~\cite{wan2013regularization}, for \emph{training} very deep networks (\eg., over 1000 layers) with stochastic depth~\cite{huang2016deep}. Unlike our method, residual layer dropping in stochastic depth networks happens only during the training stage, but at {\em test time} the layers remain fixed. Veit \etal~\cite{veit2016residual} show that ResNets are resilient to layer dropping at test time, which motivates our approach; however, they do not provide a way to dynamically choose which layers could be removed from a network without sacrificing accuracy. More recently, Huang and Wang~\cite{huang2017data} propose a method for selecting a subset of residual blocks to be executed based on a sparsity constraint. In contrast to these approaches, we propose an {\em instance-specific} residual block removal scheme to speed up ResNets during inference. %

\vspace{0.05in}
\noindent\textbf{Model Compression}. The  need to deploy top-performing deep neural network models on mobile devices motivates techniques that can effectively reduce the storage and computational costs of such networks, including knowledge distillation~\cite{Hinton15,romero2014fitnets,chenNIPS2017}, low-rank factorization \cite{ioannou2015training,tai2015convolutional,sainath2013low}, filter pruning \cite{Lecun89,polyak2015channel,li2016pruning,yu2017nisp}, quantization \cite{han2015deep,wu2016quantized,li2017training}, compression with structured matrices \cite{Circulant15,sindhwani2015structured}, network binarization \cite{rastegari2016xnor,courbariaux2016binarized,li2017sep}, and hashing \cite{chen2015compressing}. Efficient network architectures such as  {\em SqueezeNet}~\cite{SqueezeNet} and {\em MobileNet}~\cite{howard2017mobilenets} have also been explored for training compact deep nets. In contrast to this line of work where the same amount of computation is applied to all images, we focus on efficient inference by dynamically choosing a subset of blocks to be executed {\em conditioned on the input image}. More importantly, our method is {\em complementary} to these model compression techniques: the residual blocks that are kept for evaluation can be further pruned for even greater speed up.

\begin{figure*}[t!]
\begin{center}
   \includegraphics[width=1\linewidth]{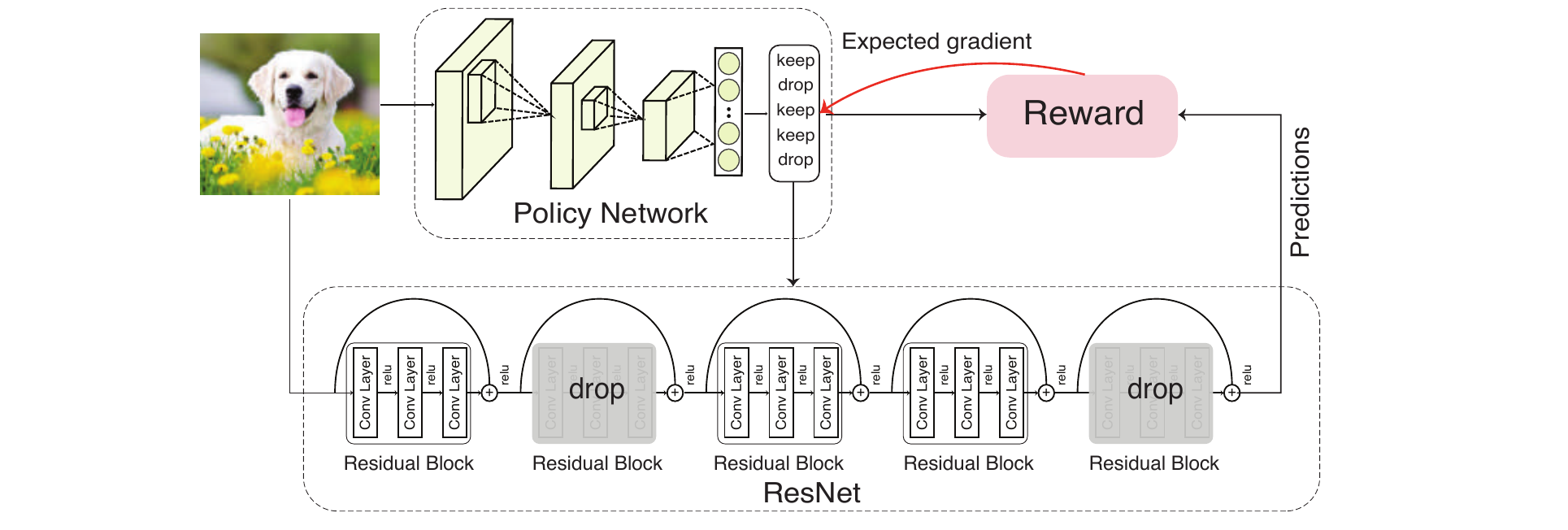}
\end{center}
\vspace{-0.3in}
   \caption{\textbf{Illustration of our proposed framework}. Given a new image, the policy network outputs dropping and keeping decisions for each block in a pretrained ResNet, which then makes a prediction by evaluating the active blocks only. Policy rewards account for both block usage and prediction accuracy. The policy network is trained to optimize the expected reward with a curriculum learning strategy, and then jointly finetuned with the ResNet.}
\label{fig:framework}
\end{figure*}
\vspace{0.05in}

\noindent\textbf{Conditional Computation}. Several {\em conditional computation} methods have been proposed to dynamically execute different modules of a network model on a per-example basis~\cite{bengio2013deep,bengio2016conditional}. Sparse activations in combination with gating functions are usually adopted to selectively turn on and off a subset of modules based on the input. These gating functions can be learned with reinforcement learning~\cite{bengio2016conditional,liu2017dynamic, denoyer2014deep}. These models typically associate a reward with a series of decisions computed after each layer/path; the resulting policy execution overhead makes it expensive to scale them up to very deep models with hundreds or thousands of layers. In contrast, our policy network makes all routing decisions in a {\em single step}, resulting in lower overhead cost for the routing itself and thus larger computational savings. %
Reinforcement learning has also been applied for dynamic feature prioritization in images~\cite{karayev2014anytime} and video~\cite{su2016leaving,yeung2016end}, actively deciding which frames or image regions to visit next. %
These techniques could be used in tandem with our approach.

\vspace{0.05in}
\noindent\textbf{Early Prediction}. Our work relates more strongly to \emph{early prediction models}, a class of conditional computation models that exit once a criterion (\eg, sufficient confidence for classification) is satisfied at early layers. Cascade detectors~\cite{felzenszwalb2010cascade,viola2004robust} are among the earliest methods that exploit this idea in computer vision, often relying on handcrafted control decisions learned separately from visual features. More recently, joint learning of features and early decisions has been studied for deep neural networks. Teerapittayanon \etal ~\cite{teerapittayanon2016branchynet} propose {\em BranchyNet}, a network composed of branches at each layer to make early classification decisions. Similarly, Adaptive Computation Time (ACT)~\cite{graves2016adaptive} augments an RNN  with a halting unit whose activation determines the probability that computation should continue.

Figurnov \etal ~\cite{figurnov2017spatially} further extend this idea to the spatial domain in ResNets by applying ACT to each spatial position of multiple image blocks.  Like our work, their formulation identifies instance-specific ResNet configurations, but it only allows configurations that use early, contiguous blocks in each predefined segment of the ResNet. These early blocks usually encode low-level features in high-dimensional feature maps, and may lack the discriminative power required for the task. This issue can be mitigated by using images at different scales~\cite{mcgill2017deciding, huang2017multi}, but at a higher computational cost. Instead, we allow \emph{any} block to contribute to our network, allowing for a much higher variability in potential ResNet configurations and policies.

\section{Approach}

Given a test image, our goal is to find the best configuration of computational blocks in a pretrained ResNet model, such that a minimum number of blocks is used, without incurring a decrease in classification accuracy. Treating the task of finding this configuration as a search problem quickly becomes intractable for deeper models as the number of potential configurations grows exponentially with the number of blocks. 
Learning a soft-attention mask over the blocks also presents problems, namely the difficulty of converting this mask into binary decisions which would require carefully handcrafted thresholds. In addition, such a thresholding operation is non-differentiable, making it non-trivial to directly adopt a supervised learning framework. 

We therefore leverage \emph{policy search methods} from reinforcement learning to derive the optimal block dropping schemes that encourage correct predictions with minimal block usage.  
To this end, we first revisit the architecture of ResNet in Sec.~\ref{sec:resnet}, and discuss why it is a good fit for block dropping. Then we introduce our  policy network in Sec.~\ref{sec:pn}, which learns to dynamically select inference paths conditioned on the input image. Finally, we present the training algorithm of our model in~Sec.~\ref{sec:training}.   
\subsection{Pretrained Residual Networks}
\label{sec:resnet}
ResNets consist of multiple stacked \emph{residual blocks} which are essentially regular convolutional layers that are bypassed by identity %
skip-connections. 
If we denote the input to the $i$-th residual block as $y_i$, and the function represented by its residual block as $\mathcal{F}_i$, the output of this residual block is given by: $y_{i+1} = \mathcal{F}_i(y_i) + y_i$, which is directly fed as input to the next residual block. 

The presence of identity skip-connections induces direct paths between any two residual blocks, and hence top layers in the network are able to access information from bottom layers during a forward pass while gradients can be directly passed from higher layers to lower layers in the back-propagation phase. Veit \etal~\cite{veit2016residual} demonstrated that removing (or dropping) a residual block at test time (\ie, having $y_{i+1} = y_i$) does not lead to a significant accuracy drop. %
This behavior is due to the fact that ResNets can be viewed as an ensemble of many paths---as opposed to single-path models like AlexNet~\cite{krizhevsky2012imagenet} and VGGNet~\cite{simonyan2014very}---and so information can be preserved even with the deletion of paths. 

The results in \cite{veit2016residual} suggest that different blocks {\em do not share strong dependencies}. However, the study also shows classification errors do increase when more blocks are removed from the model during inference.  We contend this is the result of their adopting a global dropping strategy  for all images.  We posit the best dropping schemes, which lead to correct predictions with the minimal number of blocks, must be instance-specific.

\subsection{Policy Network for Dynamic Inference Paths}
\label{sec:pn}

The \emph{configurations} in the context of ResNets represent decisions to keep/drop each block, where each decision to drop a block corresponds to removing a subset of paths from the network. We refer to these decisions as our \emph{dropping strategy}.  
To derive the optimal dropping strategy given an input instance, we develop a {\em policy network} to output a binary {\em policy vector}, representing the actions to keep or drop a block in a pretrained ResNet. During training, a reward is given considering both block usage and prediction accuracy, which is generated by running the ResNet with only active blocks in the policy vector.   See Figure~\ref{fig:framework} for an overview.

Unlike standard reinforcement learning, we train the policy to predict \emph{all actions at once}.  This is essentially a single-step Markov Decision Process (MDP) given the input state and can also be viewed as contextual bandit~\cite{langford2008epoch} or associative reinforcement learning~\cite{sutton1998reinforcement}.  %
We examine the positive impact of this design choice on scalability in Sec.~\ref{sec:quant}.

Formally, given an image ${\bf x}$ and a pretrained ResNet with $K$ residual blocks, we define a policy of block-dropping behavior as a $K$-dimensional Bernoulli distribution:
\begin{align}
\pi_{\bf W} ({\bf u}|{\bf x}) & = \prod_{k=1}^{K} {\bf s}_k^{{\bf u}_k}(1-{\bf s}_k)^{1-{\bf u}_k} \\
{\bf s}  & = f_{pn}({\bf x}; {\bf W}),
\end{align} 
where $f_{pn}$ denotes the \emph{policy network} parameterized by weights ${\bf W}$ and ${\bf s}$ is the output of the network after the \texttt{$\sigma(x)$}=$\frac{1}{1+e^{-x}}$ function. We choose the architecture of $f_{pn}$ (details below in Sec.~\ref{sec:results}) such that the cost of running it is negligible compared to ResNet, \ie, so that policy execution overhead remains low. %
The $k$-th entry of the vector, ${\bf s}^k \in [0, 1]$, represents the likelihood of its corresponding residual block in the original ResNet being dropped. An action ${\bf u}\in \{0, 1\}^{K}$ is selected based on ${\bf s}$. Here, ${\bf u}^k=0 $ and ${\bf u}^k=1 $ indicate dropping and keeping the $k$-th residual block, respectively.

Only the blocks that are not dropped according to ${\bf u}$ will be evaluated in the forward pass. To encourage both correct predictions as well as minimal block usage, we associate the actions taken with the following reward function:

\begin{align}
R({\bf u}) = 
\begin{cases}
    1- (\frac{ |{\bf u}|_0 }{K})^2   & \text{if correct}\\
    -\gamma     & \text{otherwise.} \\
\end{cases}
\label{eqn:reward}
\end{align}
Here, $(\frac{ |{\bf u}|_0 }{K})^2$ measures the percentage of blocks utilized; when a correct prediction is produced, we incentivize block dropping by giving a larger reward to a policy that uses fewer blocks. We penalize incorrect predictions with $\gamma$, which controls the trade-off between efficiency (block usage) and accuracy (\ie, a larger value leads to more correct, but less efficient policies). We use this parameter to vary the \emph{operating point} of our model, allowing different models to be trained depending on the target budget constraint.  Finally, to learn the optimal parameters of the policy network, we maximize the following expected reward:
\begin{align}
J = \mathbb{E}_{{\bf u}\thicksim {\bf \pi}_{\bf W}}[ R({\bf u})].
\label{eqn:expectation} 
\end{align}

In summary, our model works as follows: $f_{pn}$ is used to decide which blocks of the ResNet to keep conditioned on the input image, a prediction is generated by running a forward pass with the ResNet using \emph{only} these blocks, and a reward is observed based on correctness and efficiency.

\subsection{Training the BlockDrop Policy}  
\label{sec:training}

\noindent\textbf{Expected gradient}. To maximize Eqn.~\ref{eqn:expectation},  we utilize policy gradient~\cite{sutton1998reinforcement}, one of the seminal policy search methods~\cite{deisenroth2013survey}, to compute the gradients of $J$. In contrast to typical reinforcement learning methods where policies are sampled from a multinomial distribution~\cite{sutton1998reinforcement},  our policies are generated from a $K$-dimensional Bernoulli distribution. With ${\bf u}_k \in \{0,1\}$, the gradients can be derived similarly as:

\begin{align}
\nabla_{{\bf W}} J & =  \mathbb{E}[ R({\bf u})\nabla_{{\bf W}}\text{log}~\pi_{{\bf W}}({\bf u}|{\bf x})]\notag\\
&  = \mathbb{E}[ R({\bf u})\nabla_{{\bf W}} \text{log}\prod_{k=1}^{K} {\bf s}_k^{{\bf u}_k}(1-{\bf s}_k)^{1-{\bf u}_k}]\notag\\
& =  \mathbb{E}[ R({\bf u})\nabla_{{\bf W}} \sum_{k=1}^{K} \text{log} [{\bf s}_k{{\bf u}_k} + (1-{\bf s}_k)({1-{\bf u}_k})]],
\label{eqn:gradient}
\end{align}
where again ${\bf W}$ denotes the parameters of the policy network. We approximate the expected gradient in Eqn.~\ref{eqn:gradient} with Monte-Carlo sampling using all samples in a mini-batch. 
These gradient estimates are unbiased, but exhibit high variance~\cite{sutton1998reinforcement}. To reduce variance, we utilize a self-critical baseline $R(\tilde{{\bf u}})$ as in~\cite{rennie2016self} , and rewrite Eqn.~\ref{eqn:gradient} as:
\begin{align}
\label{eqn:finalgradient}
\nabla_{{\bf W}} J = ~& \mathbb{E}[A\nabla_{{\bf W}} \sum_{k=1}^{K} \text{log} [{\bf s}_k{{\bf u}_k} + (1-{\bf s}_k)({1-{\bf u}_k})]], 
\end{align}
where $A=R({\bf u})-R(\tilde{{\bf u}})$ and $\tilde{{\bf u}}$ is defined as the maximally probable configuration under the current policy, ${\bf s}$: \ie, ${\bf u}_i = 1$ if ${\bf s}_i> 0.5$, and ${\bf u}_i=0$ otherwise~\cite{rennie2016self}.

We further encourage exploration by introducing a parameter $\alpha$ to bound the distribution $\bf s$ and prevent it from saturating, by creating a modified distribution $\bf s'$:
\begin{align*}
\bf s' = \alpha \cdot \bf s + (1-\alpha) \cdot (1-\bf s). \notag
\end{align*}
This bounds the distribution in the range $1-\alpha \leq \bf s' \leq \alpha$, from which we then sample the policy vector.

\vspace{0.05in}
\noindent\textbf{Curriculum learning}. Policy gradient methods are typically extremely sensitive to their initialization.  Indeed, we found that starting from a randomly initialized policy and optimizing for both accuracy and block usage is not effective, due the extremely large dimension of the search space, which scales exponentially with the total number of blocks (there are $2^K$ possible on/off configurations of the blocks). Note that in contrast with applications such as image captioning where ground-truth action sequences (captions) can be used to train an initial policy~\cite{rennie2016self}, here no such ``expert examples" are available, other than the standard single execution path that executes all blocks.

Therefore, to efficiently search for good action sequences, we take inspiration from the idea of curriculum learning~\cite{bengio2013deep}.
During epoch $t$, for $1\le t < K$, we keep the first $K-t$ blocks on, and learn a policy only for the last $t$ blocks. As $t$ increases, the activity of more blocks are optimized, until finally all blocks are included (\ie, when $t \ge K$). Using this approach, the activation of each block is first optimized according to unmodified input features in order to assess the utility of the block, and then is gradually exposed to increasingly different feature inputs as $t$ increases and the policy for the last $t$ blocks is jointly trained. This procedure is efficient, and it is effective at identifying and removing blocks that are redundant for the input data instance being considered. \ZW{It is similar in spirit to \cite{ranzato2015sequence,rennie2016self} that gradually exposes sequences when training with REINFORCE for text generation.}

\vspace{0.05in}
\noindent\textbf{Joint finetuning}. After curriculum learning, our policy network is able to identify which residual blocks in the original ResNet to drop for a given input image. Though the policy network is trained to preserve accuracy as much as possible, removing blocks from the pre-trained ResNet will inevitably result in a mismatch between training and testing conditions. We therefore jointly finetune the ResNet with the policy network, so that it can adapt itself to the learned block dropping behavior. The principle of our joint training procedure is similar to that of stochastic depth \cite{huang2016deep}, with the exception that the drop rates are not fixed, but are instead controlled by the policy network. Alg.~\ref{alg:training} presents the complete training procedure for our framework.

\begin{algorithm}[t]
\small
  \begin{algorithmic}[1]
      \caption{The pseudo-code for training our network.}
        \label{alg:training}
    \Require{An input image ${\bf x}$ and its label}
    \State Initialize the weights of policy network {\bf W} randomly
    \State Set epochs for curriculum learning and joint finetuning to $M^{cl}$ and $M^{ft}$, respectively; and set $\alpha$ 
      \For{$t \gets 1 \textrm{ to } M^{cl} $}
        \Let {${\bf s}$}{$f_{pn}({\bf x}; {\bf W})$}
        \Let {${\bf s}$}{${\alpha \cdot \bf s + (1-\alpha) \cdot (1-\bf s)}$}
        \If{$t < K$}
          \State \texttt{set} ${\bf s}$\,{\scriptsize[1:$K-t$]}$\,=1$ \Comment{curriculum training}
        \EndIf
        \State ${\bf u} \thicksim \texttt{Bernoulli}({\bf s})$
        \State {Execute the ResNet according to} ${\bf u}$
        \State {Evaluate reward} $R({\bf u})$ {with Eqn.}~\ref{eqn:reward}
        \State Back-propagate gradients computed with Eqn.~\ref{eqn:finalgradient}
      \EndFor 
     \For{$t \gets 1 \textrm{ to } M^{ft} $}
        \State Jointly finetune ResNet and policy network
      \EndFor 
  \end{algorithmic}
\end{algorithm}

\section{Experiment}\label{sec:results}

\subsection{Experimental Setup}
\noindent\textbf{Datasets and evaluation metrics}. We evaluate our method on three benchmarks: \textsc{CIFAR-10}, \textsc{CIFAR-100}~\cite{krizhevsky2009learning}, and \textsc{ImageNet (ILSVRC2012)}~\cite{deng2009imagenet}. The CIFAR datasets consist of 60,000 32$\times$32 colored images, with 50,000 images for training and 10,000 for testing. They are labeled for 10 and 100 classes for CIFAR-10 and CIFAR-100, respectively.  Performance is measured by classification accuracy. ImageNet contains 1.2M training images labeled for 1,000 categories. We test on the validation set of 50,000 images and report top-1 accuracy.
\begin{table*}[!t]
\centering
\small
\ra{0.9}
\begin{tabular}{@{}ccccccccccccccc@{}}\toprule
& & \multicolumn{4}{c}{CIFAR-10} & & \multicolumn{4}{c}{CIFAR-100} & \\
\cmidrule{3-6} \cmidrule{8-11} 
&& Acc & $K$ & Acc ({\em ft}) & $K$ ({\em ft}) && Acc & $K$ & Acc ({\em ft}) & $K$ ({\em ft}) \\ \midrule

\multirow{5}{*}[-0.5em]{\rotatebox[origin=c]{90}{ResNet-32}} 
&FirstK & 16.6 & 10   & 84.3 & 7   &&  23.3   & 13   & 66.5    & 14  \\
&RandomK& 20.5 & 10   & 88.9 & 7   &&  38.3 & 13   & 67.6    & 14  \\
&DistributeK& 23.4   & 10  & 90.2 & 7   && 31.9   & 13  & 66.7 & 14  \\ 
&\textbf{Ours}   & \textbf{88.6} & \textbf{9.4} & \textbf{91.3} & \textbf{6.9} && \textbf{58.3} & \textbf{12.4} & \textbf{68.7} & \textbf{13.1}\\ \cmidrule{2-12}
&Full ResNet  & 92.3 & 15  & 92.3 & 15  && 69.3 & 15   & 69.3 & 15  \\ 
\midrule

\multirow{5}{*}[-0.3em]{\rotatebox[origin=c]{90}{ResNet-110}}
&FirstK & 13.3    & 21   & 71.3    & 17   && 63.5    & 50   & 57.9    & 31  \\
&RandomK& 14.5 & 21   & 90.1    & 17   && 66.3 & 50   & 68.4    & 31  \\
&DistributeK& 13.0   & 21   & 92.7  & 17   && 49.6    & 50  & 69.9 & 31  \\
&\textbf{Ours}  & \textbf{75.4} & \textbf{20.1} & \textbf{93.6} & \textbf{16.9} && \textbf{72.1} & \textbf{49.1} & \textbf{73.7} & \textbf{30.2}\\\cmidrule{2-12}
&Full ResNet  & 93.2 & 54   & 93.2 & 54   && 72.2 & 54   & 72.2 & 54\\
\bottomrule
\end{tabular}\vspace*{-0.1in}
\caption{\textbf{Accuracy and block usage with our policies vs.~heuristic baselines}, with and without jointly finetuning (ft) for all methods. For fair comparisons, $K$ is selected based on the average block usage of our method, and this can be  different before and after finetuning. Note that  the average value of $K$ for our method is reported here for brevity.  It is determined dynamically per image, and can be as low as 3 (out of 54) in ResNet-110 on CIFAR-10.
}
\vspace{-0.11in}
\label{tbl:cifar}
\end{table*}

\vspace{0.05in}
\noindent\textbf{Pretrained ResNet}. For CIFAR-10 and CIFAR-100, we experiment with two ResNet models that achieve promising results. In particular, ResNet-32 and ResNet-110 start with a convolutional layer followed by 15 and 54 residual blocks, respectively. These residual blocks, each of which contains two convolutional layers, are evenly distributed into 3 segments with down-sampling layers in between. Finally, a fully-connected layer with 10/100 neurons is applied.  See~\cite{he2015deep} for details.  For ImageNet, we adopt ResNet-101 with a total of 33 residual blocks, organized into four segments (\ie, $[3, 4, 20, 3]$). Here, each residual block contains three convolutional layers based on the {\em bottleneck} design~\cite{he2015deep} for computational efficiency.
These models are pretrained to match state-of-the-art performance on the corresponding datasets when run without our policy network.

\vspace{0.05in}
\noindent\textbf{Policy network architecture}. For our policy network, we use ResNets with a fraction of the depth of the base model. For CIFAR, we use a ResNet with 3 blocks (equivalently ResNet-8), while for ImageNet, we use a ResNet with 4 blocks (equivalently ResNet-10). In addition, we downsample images to 112$\times$112 as the input of the policy network for ImageNet experiments. The computation required for the policy network is 4.8\% and 3.0\% of the total ResNet computation for the CIFAR (ResNet-110) and ImageNet (ResNet-101) models respectively, making policy computations negligible \ZW{(it takes about 0.5 ms per image on average for ImageNet)}. While a recurrent model (\eg, LSTM) could also serve as the policy network, we found a CNN to be more efficient with similar performance.

\vspace{0.05in}
\noindent\textbf{Implementations details}. We adopt PyTorch for implementation and utilize ADAM as the optimizer.  We set $\alpha$ to 0.8, learning rate to $1e-4$, and use a batch size of 2048 during curriculum learning. 
For joint finetuning, we adjust the batch size to 256 and 320 on CIFAR and ImageNet, respectively, and  adjust the learning rate to $1e-5$ for ImageNet.  Our code is available at \url{https://goo.gl/NqyNeN}.

\vspace{0.05in}
\subsection{Quantitative Results}\label{sec:quant}
\noindent\textbf{Learned policies \vs heuristics}. We compare our block dropping strategy to the following alternative methods:
\begin{enumerate*}[label=(\arabic*)]
\item \textsc{FirstK}, which keeps only the first $K$ residual blocks active;
\item \textsc{RandomK}, which keeps $K$ randomly selected residual blocks active; 
\item \textsc{DistributeK}, which evenly distributes $K$ blocks across all segments.
\end{enumerate*} 
For all baselines, we choose $K$ to match the average number of blocks used by \system, rounding up as needed.  %
DistributeK allows us to see if feature combinations of different blocks learned by \system are better than features learned from the restricted set of early blocks of each segment. This setting resembles the allowable feature combinations from early stopping models applied to ResNets.

The results in Table~\ref{tbl:cifar} highlight the advantage of our instance-specific policy. On CIFAR-10, the learned policies give an accuracy of 88.6\% and 75.4\% using an average of 9.4 and 20.1 blocks from the original ResNet-32 and ResNet-110 respectively, outperforming the baselines by a large margin. Furthermore, the instance-specific nature of our method allows us to capture the inherent variance in the computational requirements of our dataset. We notice a wide distribution in block usage depending on the image. With ResNet-110, nearly 15\% of the images use fewer than 10 blocks, with some images using as few as 3 blocks. This variance cannot be captured by any static policies. Similar trends are observed on CIFAR-100. This confirms that dropping residual blocks with policies computed in a learned manner is indeed significantly better than heuristic dropping behaviors. 
The fact that RandomK performs better than FirstK is interesting, suggesting the value of having residual blocks at different segments to learn feature representations at different scales. %

\vspace{0.05in}
\noindent\textbf{Impact of joint finetuning}. %
Next we analyze the impact of joint finetuning (cf.~Sec.~\ref{sec:training}) for both our approach and the baselines, denoted \emph{ft} in Table~\ref{tbl:cifar}.

Joint finetuning further significantly improves classification accuracy using fewer (or almost the same) number of blocks. In particular, on CIFAR-10, it offers absolute performance gains of 2.7\% and 18.2\% using 2.5 and 3.2 {\em fewer} blocks with ResNet-32 and ResNet-110 respectively compared with curriculum training alone. Similarly, on CIFAR-100, joint finetuning improves accuracies and brings down block usage with ResNet-110. For ResNet-32, we observe 0.7 more blocks on average are used after finetuning, which might be due to the challenging nature of CIFAR-100 requiring more blocks to make correct predictions. Comparing ResNet-110 with ResNet-32, we observe that the computational speed-ups are more dramatic for deeper ResNets owing to the fact that there are more blocks with potentially diverse features to select from. 
When built upon ResNet-110, our method outperforms the pretrained model by 0.4\% and 1.5\% (absolute) using 31\% and 55.9\% of the original blocks on CIFAR-10 and CIFAR-100, respectively. Additionally, we observe that some images use as few as 5 blocks for inference. These results confirm that joint finetuning can indeed assist the ResNet to adapt to the removal of blocks by refining its feature representations while maintaining its capacity for instance-specific variation.

\vspace{0.05in}
\noindent\textbf{\system \vs state-of-the-art methods}. We next compare \system to several techniques from the literature. %
We vary $\gamma$, which controls our algorithm's trade-off between block usage and accuracy, to get a range of models with varying computational requirements. %
We compute the average FLOPs utilized to classify each image in the test set; FLOPs are a hardware independent metric, allowing for fair comparisons across models. 
\footnote{Note that we consider the multiply-accumulate operation as a two step process yielding two floating point operations and we only compute FLOPs for convolutional layers and linear layers as they account for most of the computation for inference.} 

We compare to the following state-of-the-art methods~\footnote{For ACT and SACT on CIFAR, we train models with the authors' code. For the rest, we compare to numbers in the respective papers.}: (1) ACT and (2) SACT \cite{figurnov2017spatially}, (3) PFEC \cite{li2016pruning}, (4) LCCL \cite{dong2017more}.  ACT and SACT learn a halting score at the end of each block, and exit the model when a high-confidence is obtained.
PFEC and LCCL reduce the parameters of convolutional layers by either pruning or sparsity constraints, which is complementary to our method. Other model compression methods cited earlier do not report results on larger ResNet models, and hence are not available to compare here.

\begin{figure}
\centering
\includegraphics[scale=0.74]{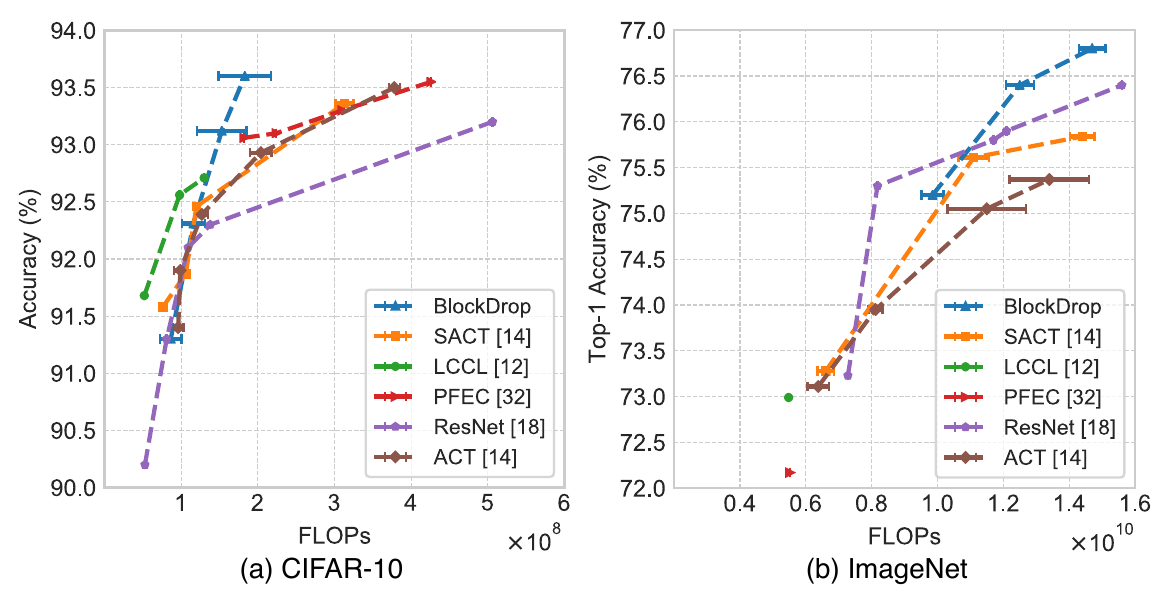}
\vspace{-0.25in}
\caption{\textbf{FLOPs \vs accuracy on CIFAR-10 and ImageNet}. Results compared to several state-of-the art methods. Error bars denote the standard deviation across images.}
\label{fig:flops}
\end{figure}

Figure~\ref{fig:flops} (a) presents the results on CIFAR. We observe that our best model offers 0.4\% performance gain in accuracy (93.6\% \vs 93.2\%) using 65\% fewer FLOPs on average ($1.73\times10^{8}$ \vs $5.08\times10^{8}$) over the original ResNet-110 model. The performance gains might result from the regularization effect of dropping blocks when finetuning the network as in~\cite{huang2016deep}.  Compared to ACT and SACT, our method only requires 50\% of the FLOPs to achieve the same level of  precision ($>$93.0\%). \system also exhibits a much higher variance in its FLOPs over other methods. Compared to SACT, this variance is 3 times larger, allowing some samples to achieve a speedup as high as 85\% with correct predictions. Further, \system also outperforms PFEC~\cite{li2016pruning} and LCCL~\cite{dong2017more}, which are complementary compression techniques and can be utilized together with our framework to speed up convolution operations. 

Figure~\ref{fig:flops} (b) presents the results for ImageNet. Compared with the original ResNet-101 model, \system again achieves slightly better results (76.8\% \vs 76.4\%) with 6\% speed up (1.47$\times10^{10}$ \vs 1.56$\times10^{10}$ FLOPs). \system performs on par with the full ResNet with a 20\% speed up (1.25$\times10^{10}$ \vs 1.56$\times10^{10}$ FLOPs) when we relax $\gamma$ slightly. This 20\% acceleration without degradation in accuracy is quite promising. For example, in a high-precision image recognition service accepting 1 billion daily API calls, such a speedup would save around 1000 hours of computation on a single P6000 GPU (0.024 seconds/image). %

\begin{figure*}[!ht]
\centering
\includegraphics[scale=0.69]{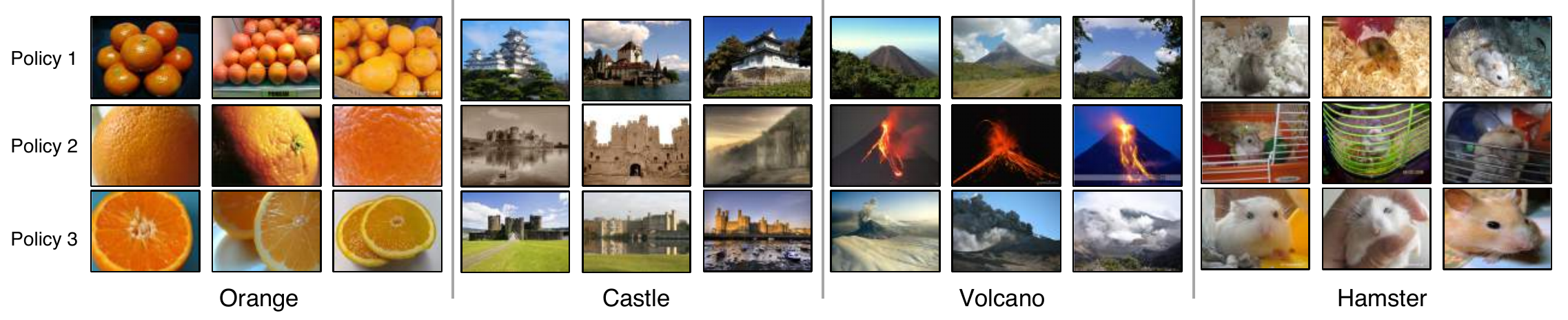}\vspace*{-0.1in}
\caption{\textbf{Policies learned for four ImageNet classes, \textit{volcano, orange, hamster} and \textit{castle}}. These policies correspond to a set of active paths in the ResNet, which seem to cater to different ``states" of images of the particular class. For \textit{volcano}, these include features like smoke, lava, \etc, while for \textit{orange} they include whether it is sliced/whole, quantity.}
\vspace{-0.2in}
\label{fig:resnet_modes}
\end{figure*}

\vspace{0.05in}
\noindent\textbf{Efficiency advantage of single-step policy}. 
\begin{table}[t]
\centering
\small
\ra{0.95}
\begin{tabular}{@{}ccccc@{}}\toprule
           &        &      Time (ms) & Speed-up \\\hline
\multirow{3}{*}[-0.12em]{\rotatebox[origin=c]{90}{{\scriptsize ResNet-32}}}   
		   & Full ResNet  & 7.71 & --      \\
           & Ours-single  &   6.56 & 14.9\%  \\
           & Ours-seq & 9.92 & -28.7\% \\
\midrule
\multirow{3}{*}[0.08em]{\rotatebox[origin=c]{90}{{\scriptsize ResNet-110}}} 
		   & Full ResNet &  24.1 & --      \\
           & Ours-single     & 10.9 & 52.3\%  \\
           & Ours-seq &  29.1 & -20.7\% \\
\bottomrule
\end{tabular}
\caption{\textbf{Impact of our single-step policy inference on efficiency for CIFAR-10}.  See text for details.}
\vspace{-0.2in}
\label{table:flops_vs_time}
\end{table}
The single-step design of our policy network---where the full  dynamic inference path is computed without revisiting intermediate outputs of the network---has important efficiency advantages.  In short, it permits lower policy execution overhead.  To examine the impact empirically, we devised a variant of \system that uses traditional RL policy learning to instead make sequential decisions (see Supp. for details).  
We select models of both variants that attain equivalent accuracy, with the same number of blocks. To ensure fair comparison, we run all three models on the same single NVIDIA P6000 GPU while disabling other processes.

Table~\ref{table:flops_vs_time} shows the results for CIFAR-10.  We report the time per test image and the speed-up over the original ResNet run in entirety with no block dropping.  
This result confirms the efficiency advantage of our single-step design: to reach the same accuracy, we need much less overhead (e.g., less than 60\% of the time required by the sequential variant).  In fact, the sequential variant takes even \emph{longer} to run than the original full ResNet models, yielding a negative speed-up.  These results reaffirm our choice to compute all actions in one shot rather than compute them sequentially.  They also stress the importance of accounting for any overhead a deep net speed-up scheme incurs to make its speed-up decisions.

\subsection{Qualitative Results}
Finally, we provide qualitative results based on our learned policies.  %
We investigate the visual patterns encoded in these learned policies and then analyze the relation between block usage and instance difficulty.

\vspace{0.05in}
\noindent\textbf{Visual patterns in policies}. Intuitively, related images can be recognized by their similar characteristics (\eg, low-level clues like texture and color). Here, we analyze similarity in terms of the \emph{policies they utilize} by sampling dominant policies for each class and visualizing samples from them. 
Figure~\ref{fig:resnet_modes} shows samples utilizing three different policies for four classes. It can be clearly seen that images under the same policy are similar, and different policies encode different styles, although they all correspond to the same semantic concept. For example, the first inference path for the ``orange'' class caters to images containing a pile of oranges, and close up views of oranges activate the second inference path, while images containing slices of oranges are routed through the third inference path. These results indicate that different paths encode meaningful semantic visual patterns, based on the input images. While this happens in standard ResNets as well, all images necessarily utilize all the paths, and disentangling this information is not possible.

\vspace{0.05in}
\noindent\textbf{Instance difficulty}.
Instance difficulty is well understood in the context of prediction confidence, where easy and difficult examples are classified with high and low probabilities, respectively. 
Inspired by the above analysis that revealed interesting correlations between the inference policies and the visual patterns in the images, 
we try to characterize instance difficulty in terms of block usage. We hypothesize that simple examples (\eg images with clear objects, without occlusions) require fewer computations to be correctly recognized. 
To qualitatively analyze the correlations between instance difficulty and block usage, we utilize learned policies that lead to high-confidence predictions for each class. %

\begin{figure}[h!]
\centering
\includegraphics[width=0.92\linewidth]{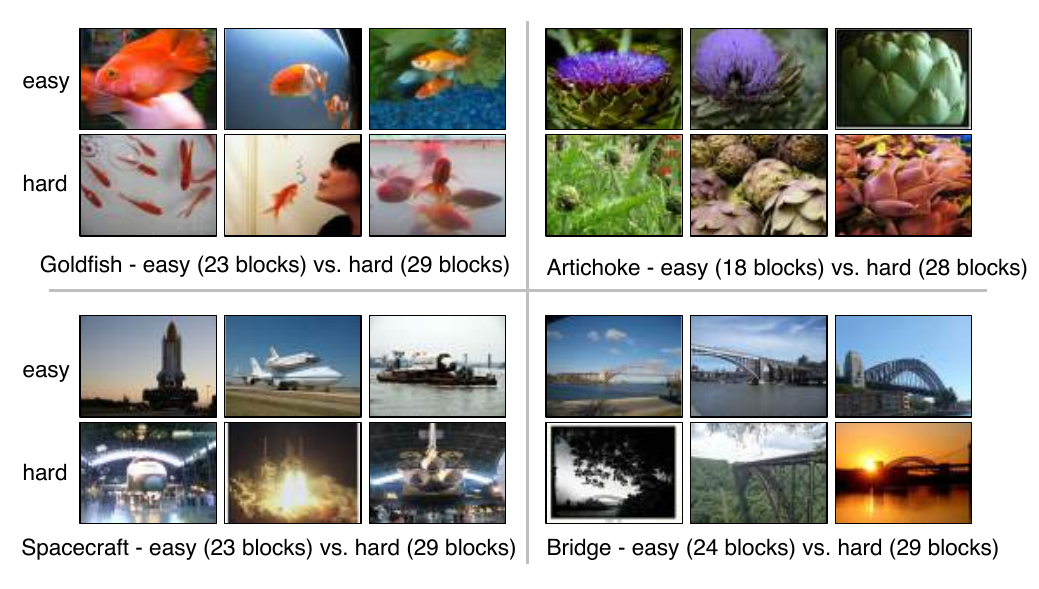}
\vspace{-0.15in}
\caption{\textbf{Samples from ImageNet classes}. Easy and hard samples from \textit{goldfish}, \textit{artichoke}, \textit{spacecraft} and \textit{bridge} to illustrate how block usage translates to instance difficulty.}
\vspace{-0.26in}
\label{fig:difficulty}
\end{figure}

Figure~\ref{fig:difficulty} illustrates samples from ImageNet. The top row contains images that are correctly classified with the least number of blocks, while samples in the bottom row utilize the most blocks. We see that samples using fewer blocks are indeed easier to identify since they contain single frontal-view objects positioned in the center, while several objects, occlusion, or cluttered background occur in samples that require more blocks. This confirms our hypothesis that block usage is a function of instance difficulty.  We stress that this ``sorting" into easy or hard cases falls out automatically; it is learned by \system.

\vspace{-0.1in}
\section{Conclusion}
\vspace{-0.05in}
We presented \system, an approach for faster inference in ResNets by selectively choosing residual blocks to evaluate in a learned and optimized manner conditioned on inputs. In particular, we trained a policy network to predict blocks to drop in a pretrained ResNet while trying to retain the prediction accuracy.
The ResNet is further jointly finetuned to produce smooth feature representations tailored for block dropping behavior. We conducted extensive experiments on CIFAR and ImageNet, observing considerable gains over existing methods in terms of the efficiency-accuracy trade-off. Further, we also observe that the policies learned encode semantic information in the images.  

{\scriptsize \noindent\textbf{Acknowledgments}: Rogerio Feris is supported by IARPA via DOI/IBC contract number D17PC00341. The U.S. Government is authorized to reproduce and distribute reprints for Governmental purposes notwithstanding any copyright annotation thereon. Disclaimer: The views and conclusions contained herein are those of the authors and should not be interpreted as necessarily representing the official policies or endorsements, either expressed or implied, of IARPA, DOI/IBC, or the U.S. Government. Kristen Grauman is supported in part by an IBM Faculty Award and IBM Open Collaboration Award. Larry S. Davis is partially supported by the Office of Naval Research under Grant N000141612713.}

{\small
\bibliographystyle{ieee}
\bibliography{egbib}
}

\clearpage
\onecolumn
{\centering
\section*{Supplemental Materials}}

\subsection*{Details of BlockDrop-seq (Ours-seq)}
We construct a sequential version of \system for dropping blocks, where the decision ${\bf a}_i \in \{0,1\}$ to drop or keep the $i$-th block is conditioned on the activations of its previous block, $y_{i-1}$. Unlike \system, where all the actions are predicted in one shot, this model predicts one action at a time, which is a typical reinforcement learning setting. We follow the procedure to generate the \emph{halting scores} in \cite{figurnov2017spatially}, and arrive at an equivalent per-block \emph{skipping score}
according to:
\begin{align*}
{\bf p}_i = \texttt{softmax}(\widetilde{W}^i \texttt{pool}(y_{i-1}) + b^i),
\end{align*}
where $\texttt{pool}$ is a global average pooling operation.
For fair comparisons, Ours-seq is compared to a \system model, which attains equivalent accuracy, with the same number of blocks.

\subsection*{Implementation Details}
\begin{itemize}

  \item On CIFAR, we train the model for 5000 epochs during curriculum learning with a batch size of 2048 and a learning rate of $1e-4$. We further jointly finetune the model for 1600 epochs with a batch size of 256 and a learning rate of $1e-4$, which is annealed to $1e-5$ for 400 epochs.
  
  \item  On ImageNet, the policy network is trained for 45 epochs for curriculum learning with a batch size of 2048 and a learning rate of $1e-4$. We then use a batch size of 320 during joint finetuning for 10 epochs.
  
\end{itemize} 

\subsection*{Detailed Results on CIFAR-10 and ImageNet}
We present detailed results of our method on CIFAR-10 (Table~\ref{table:cifar_results}) and ImageNet (Table~\ref{table:imagenet_results}). We highlight the accuracy, block usage and speed up for variants of our model compared to full ResNets. 

\begin{table*}[!h]
\centering
\begin{tabular}{@{}ccccccc@{}}\toprule
   &    Network               & FLOPs                & Block Usage  & Accuracy & Speed-up \\
\cmidrule{1-6}
 & ResNet-32 & 1.38E+08 $\pm$ 0.00E+00 & 15.0 $\pm$ 0.0 & 92.3     & --       \\
 & ResNet-110          & 5.06E+08 $\pm$ 0.00E+00 & 54.0 $\pm$ 0.0 & 93.2     & --       \\

\cmidrule{1-6} 
& \system-32 ($\gamma=5$)  & 8.66E+07 $\pm$ 1.40E+07 & 6.9 $\pm$ 1.6  & 91.3     & 37.2\%   \\
& \system-110 ($\gamma=2$)  & 1.18E+08 $\pm$ 2.46E+07 & 10.3 $\pm$ 2.7 & 91.9     & 76.7\%   \\
& \system-110 ($\gamma=5$)  & 1.51E+08 $\pm$ 3.24E+07 & 13.8 $\pm$ 3.5 & 93.0     & 70.1\%   \\
& \system-110 ($\gamma=10$) & 1.81E+08 $\pm$ 3.43E+07 & 16.9 $\pm$ 3.7 & 93.6     & 64.3\%   \\
\bottomrule
\end{tabular}
\caption{Results of different architectures on CIFAR-10. Depending on the base ResNet architecture, speedups ranging from 37\% to 76\% are observed with little to no degradation in performance. }
\label{table:cifar_results}
\end{table*}

\begin{table*}[h!]
\centering
\begin{tabular}{@{}ccccccc@{}}\toprule
& Network & FLOPs & Block Usage & Accuracy & Speed-up \\
\cmidrule{1-6}
& ResNet-72 & 1.17E+10 $\pm$ 0.00E+00 & 24.0 $\pm$ 0.0 & 75.8  &  -- \\
& ResNet-75 & 1.21E+10 $\pm$ 0.00E+00 & 25.0 $\pm$ 0.0 & 75.9  &  -- \\
& ResNet-84 & 1.34E+10 $\pm$ 0.00E+00 & 28.0 $\pm$ 0.0 & 76.1  &  -- \\
\cmidrule{1-6}
& ResNet-101          & 1.56E+10 $\pm$ 0.00E+00 & 33.0 $\pm$ 0.0 & 76.4     & --       \\
\cmidrule{1-6}
& \system ($\gamma=2$) & 9.85E+09 $\pm$ 3.34E+08 & 18.8 $\pm$ 0.8 & 75.2     & 36.9\%   \\
& \system ($\gamma=5$)  & 1.25E+10 $\pm$ 4.26E+08 & 24.8 $\pm$ 1.0 & 76.4     & 19.9\%   \\
& \system ($\gamma=10$) & 1.47E+10 $\pm$ 4.02E+08 & 29.7 $\pm$ 0.9 & 76.8     & 5.7\%    \\
\bottomrule
\end{tabular}
\caption{Results of different architectures on ImageNet. \system is built upon ResNet-101, and can achieve around 20\% speedup on average with $\gamma=5$.}
\label{table:imagenet_results}
\end{table*}

\end{document}